\AtBeginDvi{}
\RequirePackage{plautopatch}

\documentclass[english, techrep]{ipsj}

\makeatletter
\newcommand\newblock{\hskip .11em\@plus.33em\@minus.07em}
\makeatother

\usepackage[whole]{bxcjkjatype} 
\usepackage[unicode]{hyperref} 

\usepackage{autobreak}
\usepackage{algorithm}
\usepackage{amsmath}
\usepackage[noend]{algpseudocode}
\usepackage{bm}
\usepackage{bmpsize}
\usepackage{bbm}
\usepackage{cite}
\usepackage{float}
\usepackage[pdftex]{graphicx}
\usepackage{latexsym}
\usepackage{textcomp}
\usepackage{times}
\usepackage{xcolor}
\usepackage{comment}
\usepackage{url}

\def\Underline{\setbox0\hbox\bgroup\let\\\endUnderline}
\def\endUnderline{\vphantom{y}\egroup\smash{\underline{\box0}}\\}
\def\|{\verb|}

\makeatletter
\def\@uketsuke{}
\def\@euketsuke{}
\def\SHUBETUname{}
\makeatother

\AtBeginDocument{
 \abovedisplayskip     =0.5\abovedisplayskip
 \abovedisplayshortskip=0.5\abovedisplayshortskip
 \belowdisplayskip     =0.5\belowdisplayskip
 \belowdisplayshortskip=0.5\belowdisplayshortskip
}

\algnewcommand\algorithmicinput{{\bfseries\gtfamily Input:}}%
\algnewcommand\algorithmicoutput{{\bfseries\gtfamily Output:}}%
\algnewcommand\AlgInput{\item[\algorithmicinput]}%
\algnewcommand\AlgOutput{\item[\algorithmicoutput]}%
\algrenewcommand\Return{\State\textbf{return} }%

\begin{document}

\title{Estimating the number of reachable positions in Minishogi}
\affiliate{utokyo-c}{Graduate School of Arts and Sciences, The University of Tokyo}
\affiliate{utokyo-itc}{Information Technology Center, The University of Tokyo}

\author{Sotaro Ishii}{utokyo-c}
\author{Tetsuro Tanaka}{utokyo-itc}

\begin{abstract}
To investigate the feasibility of strongly solving Minishogi (Gogo Shogi), it is necessary to know the number of its reachable positions from the initial position. However, there currently remains a significant gap between the lower and upper bounds of the value, since checking the legality of a Minishogi position is difficult. In this paper, the authors estimate the number of reachable positions by generating candidate positions using uniform random sampling and measuring the proportion of those reachable by a series of legal moves from the initial position. The experimental results reveal that the number of reachable Minishogi positions is approximately $2.38\times 10^{18}$.
\end{abstract}

\maketitle

\renewcommand{\thefootnote}{\fnsymbol{footnote}}
\footnotetext[1]{This paper is a modified version of ``Estimating the number of reachable positions in Minishogi.'' (IPSJ SIG Technical Reports, Vol. 2024-GI-53, No.2, pp.1-6)}
\renewcommand{\thefootnote}{\arabic{footnote}}
\setcounter{footnote}{0}

\section{Introduction}
Among board games, Shogi, Chess, and Go are theoretically classified as two-player, zero-sum, finite, deterministic, perfect information games. In these games, the game value (win, loss, or draw) can be determined for every possible position. The levels of solving games are classified into the following three categories\cite{allis-1994}.

\begin{itemize}
   \item Ultra-weakly solved: The game value of the initial position is known, but the best moves are not.
   \item Weakly solved: The game value of the initial position is known, and the best moves are also known for the positions necessary to calculate the value of the initial position.
   \item Strongly solved: The game value and the best moves are known for all positions reachable from the initial position.
\end{itemize}

Strongly solving a game can be achieved either by enumerating all positions reachable from the initial position or by performing retrograde analysis, which searches from the terminal positions back to the initial position. However, these methods are not applicable to large-scale games in practice. Therefore, it is essential to understand the scale of a game when considering the feasibility of a strong solution. One important metric for assessing the ``scale'' is the number of positions reachable from the initial position, known as the ``state-space complexity'' of the game\cite{allis-1994} (referred to as ``the number of reachable positions'' in this study).

Previous studies on the number of reachable positions in various games typically used approaches that either precisely calculate the number of board configurations or strictly evaluate the upper and lower bounds. In contrast, we applied a statistical estimation method to approximate the number of reachable positions in Minishogi. Specifically, following prior research in chess, we generated a large number of candidate positions (hereafter referred to as ``pseudo-legal positions'') using random numbers and applied an algorithm that determines the reachability from the initial position through legal moves in reverse order to estimate the number of reachable positions.

By leveraging the feature of Minishogi, where pieces can be dropped onto the board, we efficiently determined the reachability by performing a best-first search with a heuristic function, targeting positions where the Manhattan distance between both kings is more than two.

\section{Minishogi}
\begin{figure}[tb]
 \begin{center}
 \begin{minipage}[c]{0.6\columnwidth}
   \centering
   \includegraphics[width=\columnwidth]{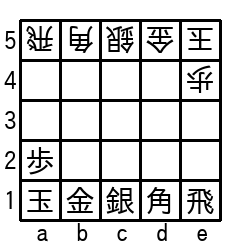}
 \end{minipage}
 \end{center}
 \caption{The initial position of Minishogi}
 \label{init-pos}
\end{figure}

Minishogi (Gogo Shogi) is a variant of Shogi played on a 5x5 grid, which is smaller than standard Shogi. It was invented by Shigenobu Kusumoto in 1970, and the game starts from the initial position shown in \figref{init-pos}, with players alternating moves\cite{umebayashi}. The goal of the game, as in standard Shogi, is to checkmate the opponent's king. The rules of Minishogi generally adhere to those of standard Shogi, with differences in board size and the number of pieces used, as follows:

\begin{itemize}
   \item The game uses six types of pieces: King (玉), Gold (金), Silver (銀), Bishop (角), Rook (飛), and Pawn (歩).
   \item Captured enemy pieces become part of the player's own pieces (in-hand pieces) and can be dropped onto any vacant square on the board, provided no foul move is committed.
   \item Silver, Bishop, Rook, and Pawn can promote to Promoted Silver (全), Horse (馬), Dragon (竜), and Promoted Pawn (と), respectively, when they enter, leave or move within the enemy camp. However, they can't be promoted when they are dropped.
   \item When the condition for promotion is met, the player can choose whether to promote the piece or not. However, if the piece would become immobile by not promoting, promotion is mandatory.
   \item The following moves are fouls:
       \begin{itemize}
           \item Two Pawns: Having two or more of the player's Pawns on the same file.
           \item Immobile Piece: Creating a situation where a piece cannot be moved at all until the end of the game.
           \item Drop Pawn Mate: Dropping a Pawn to checkmate the opponent's King.
           \item Ignoring a check on the player's own King.
       \end{itemize}
\end{itemize}

Additionally, in actual games, it is common to add a rule that ``repetitions (Sennichite) result in a win for the second player'' to actively encourage the first player to avoid such situations \cite{mizuta}.

In recent years, research on Minishogi has progressed with the development of strong AIs, such as those developed by Shioda et al.\cite{shioda}, leading to the proof of a win for the second player in some initial moves\cite{mizuta}. Due to these developments, Minishogi is expected to be weakly solved in the near future; however, the feasibility of the strong solution has not yet been concretely discussed. One reason for this is that the exact number of reachable positions in Minishogi remains unknown. Previous research\cite{miyako} has shown that the number lies between $4.59 \times 10^{12}$ and $1.95 \times 10^{19}$, but this range is too large to determine the feasibility of a strong solution. This study aims to estimate the number of reachable positions in Minishogi directly and evaluate the feasibility of a strong solution.

In this study, a position is defined as a combination of the board configuration, in-hand pieces, and the player's turn while the history of moves from the initial position or the number of consecutive checks is not considered. If the goal of the strong solution is to determine the game results after any legal sequence of moves from the initial position, then the answer might differ depending on whether a position has appeared once or three times. However, in order to determine the game result without any position appearing more than once from the initial position, our definition of the position is sufficient.

Furthermore, to consider positions in the second player's turn, it is sufficient to flip the board of a position in the first player's turn by 180 degrees. Therefore, we restricted the player's turn to the first player. Although it might seem appropriate to treat positions with and without repetition differently in order to determine the game result without considering draws under the current Sennichite rule, it is not problematic because a strong solution with a three-valued game result (win, loss, draw) under the Sennichite draw rule can be extended to handle cases without considering draws. Additionally, all piece movements in Minishogi are horizontally symmetrical, so we treated horizontally flipped positions as identical.

Note that previous studies\cite{miyako, shinoda} have defined a position as a combination of the board configuration, in-hand pieces, and the player's turn, treating positions that are vertically flipped as identical but distinguishing horizontally flipped positions. Following this definition, the estimated number of reachable positions in this study is approximately half of those in previous studies. Depending on the game's rules, the reachability from the initial position may differ in horizontally flipped positions. However, in Minishogi, the position obtained by horizontally flipping the initial position is also reachable from the initial position. This fact indicates that if a position is reachable from the initial position, the horizontally flipped position is also reachable.

\section{Related Research}
In this section, we introduce previous research on estimating the number of reachable positions in various games. Allis\cite{allis-1994, allis-1988} has estimated the number of reachable positions in games like Connect Four and Tic-Tac-Toe. Since these are games where the pieces placed do not move or get removed, the number of reachable positions can be easily determined by exploring from the initial position or calculating the number of board configurations. In Chess, Shannon\cite{shannon} provided a famous estimate of $10^{42}$, and Chinchalkar\cite{chinchalkar} calculated an upper bound of $1.78 \times 10^{46}$.

In Shogi, Shinoda\cite{shinoda} estimated the number of reachable positions in Shogi to be between $4.65 \times 10^{62}$ and $9.14 \times 10^{69}$ by counting positions that do not violate the rules. Miyako et al.\cite{miyako} improved this evaluation and derived a range of $2.45 \times 10^{64}$ to $6.78 \times 10^{69}$, and also calculated the upper and lower bounds of the number of positions in Minishogi using the same method.

Some previous studies determined the number of reachable positions using different approaches. Takeda et al.\cite{takeda} constructed a minimal perfect hash function for ``Nine Men's Morris''-type stone-capturing games and calculated their number of reachable positions. Tromp\cite{tromp-cpr} proposed a method of statistically estimating the number of reachable positions by generating a large number of pseudo-legal positions randomly and counting the ones that are reachable from the initial position. Besides, Tromp applied this method to chess and estimated the number of reachable positions to be $(4.82 \pm 0.03) \times 10^{44}$ based on the sampling of 2 million positions \cite{tromp-cpr}. Our study can be considered an application of Tromp's method to Minishogi.

\section{Sampling based Estimation}
This section describes the method for estimating the number of reachable positions in Minishogi. In games like Connect Four, Tic-Tac-Toe, or Go, it is relatively easy to determine the number of reachable positions as the number of legal board configurations, as there are fewer constraints on the legality of moves. However, in chess-like games including Shogi, accurately counting board configurations that are ``reachable only through legal moves from the initial position'' is not obvious. In Dobutsu Shogi, another simplified variant of Shogi, the game is decided by capturing the king, while the game of Minishogi is decided when there are no legal moves to escape a check. This fact makes determining reachability from the initial position more difficult in Minishogi, as well as the standard Shogi.

First, we generated the candidate set $S_{all}$ containing all possible positions and assigned a unique rank between 0 and $|S_{all}|-1$ to each element. Then we randomly extracted a sufficient number of positions from $S_{all}$ with equal probability and estimated the probability $p$ by checking whether these positions are reachable or not. Finally we estimated the expected number of reachable positions as $p \cdot |S_{all}|$. 

Tromp\cite{tromp-cpr} and we both estimate the number of states from random sampling of pseudo-legal positions, but our study differs in that it determines the legality of a position based on its reachability to a specific position.

\subsection{Generation of Candidate Positions Set}
The elements of $S_{all}$ are positions that meet the following conditions:

\begin{itemize}
\item Fix the player's turn to the first player, and do not generate positions in the second player's turn.
\item Restrict the location of the first player's king to files a-c. When the first player's king is in file c, restrict the second player's king to files a-c.\footnote{This restriction aims to reduce the generation of positions that are identical when flipped horizontally.}
\item Place the pieces on the board without any constraints on the empty squares.
\end{itemize}

The generation procedure is as follows:

\begin{enumerate}
   \item\label{piece-enum-1} Enumerate all possible patterns for each piece except the king (Gold, Silver, Bishop, Rook, Pawn) for how many of each piece can exist on the board and in the hand (without distinguishing between the players) (Promoted pieces are treated as identical to the unpromoted pieces).
  
   The pattern generated here is represented as a tuple $(hc, bc)$, where $hc$ denotes a list specifying the quantity and type of pieces in players' hands, and $bc$ represents a list detailing the quantity and type of pieces on the board. An example of such a tuple could be [(Gold, 1), (Pawn, 1)] for $hc$, [(Gold, 1), (Pawn, 1), (Silver, 2), (Rook, 2), (Bishop, 2)] for $bc$. This pattern represents the following situation:
   \begin{itemize}
       \item Pieces off the board (without distinguishing between players) are one Gold and one Pawn.
       \item Pieces on the board are one Gold, one Pawn, two Silvers, two Rooks, and two Bishops.
   \end{itemize}
   \item\label{piece-enum-2} For the patterns generated in \ref{piece-enum-1}, calculate the total number of possible positions $N_c$ corresponding to each pattern $c$ using the functions $C$ and count2N described below.
   \item Generate positions based on the information $(c, N_c)$ and rank them using the value of $N_c$.
\end{enumerate}

The total number of ways to place $v$ pieces of type $p$ on $n_\text{empty}$ empty squares on the board is given by the following function $C(p, n_{empty}, v)$:

\begin{align*}
C\left(p, n_\text{empty}, v \right) &= \sum_{p_{0}=0}^{v \cdot \mathbbm{1}(p)} \ \sum_{p_{1}=0}^{(v - p_0) \cdot \mathbbm{1}(p)} \ \sum_{\bar{p}_{0}=0}^{v - p_{0} - p_{1}} \\
&\binom{n_\text{empty}}{p_0} \cdot \binom{n_\text{empty} - p_0}{p_1} \\
&\cdot \binom{n_\text{empty} - p_0 - p_1}{\bar{p}_0} \\
&\cdot \binom{n_\text{empty} - p_0 - p_1 - \bar{p}_0}{v - p_0 - p_1 - \bar{p}_0}
\end{align*}

Here,
\begin{itemize}
   \item $p_0, \bar{p}_0$: The number of promoted and unpromoted pieces of the first player
   \item $p_1$: The number of promoted pieces of the second player
   \item $\mathbbm{1}(p) = \begin{cases}
                       1 & (\text{If} \ p \in \{ \mathrm{Silver, Rook, Bishop, Pawn} \})\\
                       0 & (\text{If} \ p \in \{ \mathrm{King, Gold} \})
                   \end{cases}$
\end{itemize}

Using this function $C(p, n_{empty}, v)$, the function count2N (Algorithm \ref{count2N}) calculates the total number $N_c$ of positions corresponding to the combination $c = (hc, bc)$. The sum of the calculated $N_c$ values for each $c$ gives the total number of candidate positions $|S_{all}|$.

\begin{algorithm}
\caption{count2N}
\label{count2N}
\begin{algorithmic}[1]
\AlgInput{Tuple $c = (\text{hc}, \text{bc})$}
\AlgOutput{$N_c$: Total number of positions}
\State $N_{hc} \gets 1$
\ForAll {$(p, v) \in \text{hc}$} \Comment{Calculation of the number of combinations for in-hand pieces}
   \State $N_{hc} \gets N_{hc} \cdot (v + 1)$
\EndFor
\State $N_{bc} \gets \text{KPOS\_COUNT}$ \Comment{Number of combinations of king positions}

\State $n_\text{empty} \gets 5 \times 5 - 2$ \Comment{Number of empty squares excluding the kings (23 squares)}
\ForAll {$(p, v) \in \text{bc}$} \Comment{Calculation of the number of combinations for board pieces}
   \State $x \gets C(p, n_\text{empty}, v)$
   \State $N_{bc} \gets N_{bc} \cdot x$

   \State $n_\text{empty} \gets n_\text{empty} - v$
\EndFor

\Return $N_{hc} \cdot N_{bc}$
\end{algorithmic}
\end{algorithm}

Here, $\text{KPOS\_COUNT}$ is the number of combinations of king positions, calculated from the number of ranks (rows)  $H = 5$ and files (columns) $W=5$ on the board as follows:
\begin{align*}
\text{KPOS\_COUNT} &= H \times \left\lfloor\frac{W}{2}\right\rfloor \times \left(H \times W - 1\right) \\
&+ H \times \left(H \times \left\lfloor\frac{W+1}{2}\right\rfloor -1\right) \\
&= 310   
\end{align*}
The first term represents the case where the first player's king is placed in files a-b, and the second term represents the case where the first player's king is placed in file c.

The total number of candidate positions in $S_{all}$ generated by the above procedure is $16,014,219,505,238,849,250$ $\left( \approx 1.6 \times 10^{19} \right)$. The integer rank between 0 and $|S_{all}|-1$ and the position (combination of board configuration, in-hand pieces, and player's turn) can be converted to each other. We omit the details of the conversion in this paper.

\subsection{Determining the Legality of Generated Positions}
We eliminated illegal positions from $S_{all}$, the generated pseudo-legal positions, according to the following steps.

\begin{figure}[tb]
\begin{center}
   \includegraphics[width=4cm]{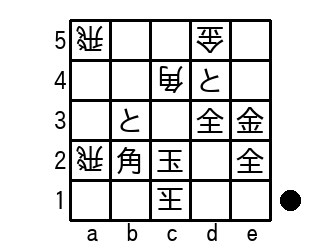}
   \includegraphics[width=4cm]{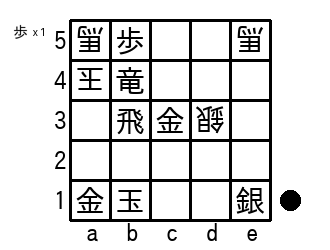}\\
   \includegraphics[width=4cm]{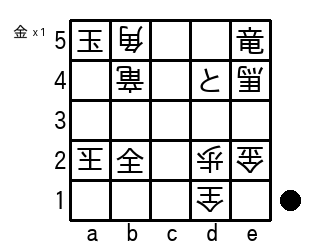}
   \includegraphics[width=4cm]{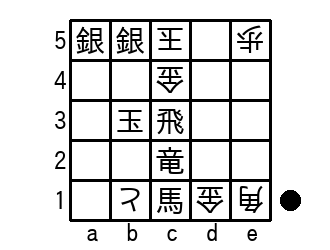}
\end{center}
\caption{Unreachable positions.}
\label{impossible-ex1}
\end{figure}

\begin{enumerate}
\item Eliminate one of the positions that are identical when flipped horizontally, keeping the one with the smaller lexicographic order. In \figref{impossible-ex1}, the upper left position has no order relationship regarding the kings when flipped horizontally. However, the first player's Gold is located on e3, which is higher in lexicographic order than its flipped position a3, so the flipped position is smaller.
\item Eliminate positions that contain the Two Pawns or an immobile pawn. In the upper right position of \figref{impossible-ex1}, the first player's pawn in b5 is immobile.
\item Eliminate positions where the second player's king is in check despite it being the first player's turn. In the lower left position of \figref{impossible-ex1}, the first player's promoted Silver in b2 is checking the second player's king in a2.
\item\label{cant_reach} Eliminate positions that cannot be traced back to the initial position. Most positions that fall into this category are those where the first player's king is in double check and cannot be traced back even one move. However, there are also cases like the one in the lower right of \figref{impossible-ex1}, where it is possible to trace back one move but not two (assuming the previous move was the second player dropping a Gold on c4).
\end{enumerate}

To detect positions that fall under step \ref{cant_reach}, we defined the following heuristic function $H(pos)$ for a position $pos$.
\[
H(pos) = a \cdot N(pos) + b \cdot P(pos) + c \cdot D(pos) + d \cdot K(pos)
\]

Here,

\begin{itemize}
   \setlength{\itemsep}{1pt}
   \item ${a, b, c, d}$: Real-valued parameters
   \item $N(pos)$: The number of pieces on the board other than the kings
   \item $P(pos)$: The number of promoted pieces on the board
   \item $D(pos)$: The total number of squares that all promoted pieces are away from the first row of the enemy's camp in terms of vertical distance\footnote{The more pieces that have returned to the player's camp after promoting in the enemy's camp, the less likely it is that the position is reachable.} (If there are no promoted pieces, then $0$)
   \item $K(pos) = \begin{cases}
                       1 & (d_{\mathrm{KINGS}} \leq 2) \\
                       0 & (\text{otherwise})
                   \end{cases}$\footnote{Positions where the Manhattan distance between the kings is two or less include situations where the kings are checking each other. These cannot be reached without self-check, so they are illegal.} \\
                   where $d_{\mathrm{KINGS}}$ is the Manhattan distance between the kings.
\end{itemize}

A $pos$ that satisfies $H(pos) = 0$ is a position where only the two kings are present on the board, and the Manhattan distance between them is more than two (hereafter referred to as ``KK positions''). KK positions can be reached from the initial position by both players capturing all of the opponent's pieces except for the kings. Also, since a player facing a KK position can transfer his in-hand pieces to the opponent's hand without changing the positions of the kings and the player's turn, any position that can trace back to a KK position can also trace back to the initial position through the KK position. In this study, we used the parameters $a = 10, b = 10, c = 1, d = 1$.

To determine the reachability of a position $pos$, we performed a greedy best-first search using the function $prev(pos)$, which generates the set of legal positions one move before $pos$ (Algorithm \ref{greedy-bfs}).

\begin{algorithm}[H]
   \caption{can\_reach\_KK}
   \label{greedy-bfs}
   \begin{algorithmic}[1]
       \AlgInput{Position $pos$}
       \AlgOutput{Succeeded or Failed}
       \State $\bm{q} \gets \{pos\}$
       \While{$\bm{q}$ is not Empty}
           \State $pos1 \gets$ \textit{pos} s.t. $\underset{pos \in \bm{q}} {\operatorname{argmin}} \ H(pos)$
           \State $\bm{q} \gets \bm{q} - \{pos1\}$
           \If{$H(pos1) = 0$}
              \Return Succeeded
           \EndIf
           \ForAll{$pos2 \in prev(pos1)$}
               \State $\bm{q} \gets \bm{q} \cup \{pos2\}$
           \EndFor
       \EndWhile
       \Return Failed
   \end{algorithmic}
\end{algorithm}

When the search diverges, this program could run out of memory and terminate abnormally, but so far, it has produced results in a practical time for all the positions tested.

\section{Experiment}
We generated 100,000,000 unique random numbers in the range 0 to $|S_{all}| - 1$, obtained their corresponding positions, and counted how many of these positions passed the checks described in the previous section. The program was implemented in Python for an emphasis on readability and rapid prototyping. The calculations were performed in parallel on a 128GB AMD Ryzen Threadripper 2990WX machine over about three days. As a result, 14,849,198 positions were identified as reachable. The number of positions that passed each check is shown in \tabref{counts}.

\begin{table}[tb]
\caption{Number of positions that passed the tests}
\label{counts}
\hbox to\hsize{\hfil
\scalebox{1.2}{
   \begin{tabular}{|l|r|} \hline
       Check Content & Number Passed \\ \hline \hline
       Initial Generation & 100,000,000 \\ \hline
       Horizontal Flip & 96,774,076 \\ \hline
       Pawn Placement & 77,795,825  \\ \hline
       Opponent King Check & 21,506,911 \\ \hline
       Reachability & 14,849,198 \\ \hline
   \end{tabular}
}
\hfil}
\end{table}

Based on these results, the probability $p$ is estimated to be between $0.14842\ldots$ and $0.14856\ldots$ with a $95\%$ confidence interval. Using this estimate, the expected number of reachable positions $|S_{ok}|$ is calculated to be between $2.376\ldots \times 10^{18}$ and $2.379\ldots \times 10^{18}$. Therefore, the number of reachable Minishogi positions is estimated to be $2.38 \times 10^{18}$ with three significant digits.

\begin{figure*}[tb]
\begin{center}
   \includegraphics[width=16cm]{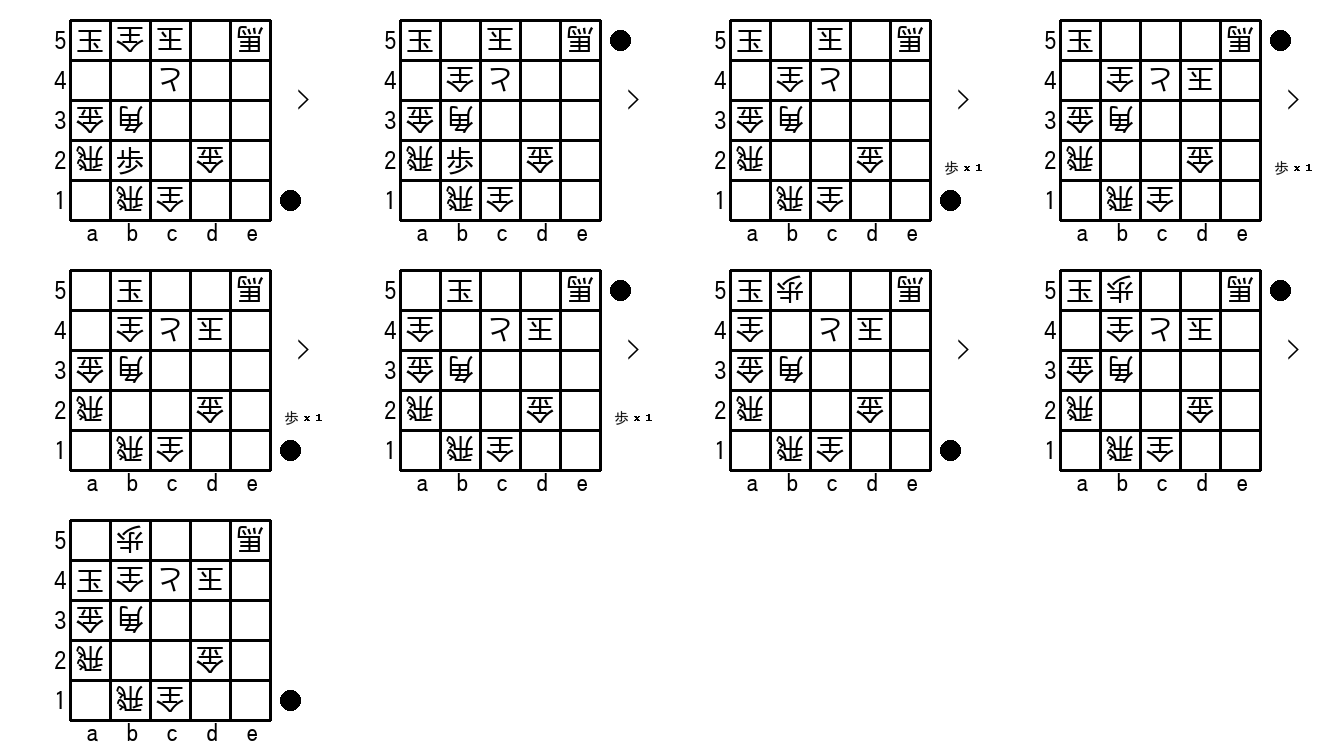}
\end{center}
\caption{The unreachable position that can be traced back 8 plies}
\label{long_noreach}
\end{figure*}

For positions that did not pass the reachability check, \tabref{count_cant_reach} shows the distribution of the maximum number of moves that could be traced back. The position that could be traced back the longest, up to 8 moves, is shown in \figref{long_noreach}. There may be unreachable positions that can be traced back more moves, but determining the maximum number of moves remains unresolved. Ensuring that the search does not diverge is a crucial factor for the applicability of this method to Minishogi. It is observed that Minishogi has the property that the search from unreachable positions stops in a small number of moves, and the search from reachable positions proceeds in the direction of the steepest descent of the heuristic function without much backtracking.

\begin{table}[tb]
\caption{Number of the unreachable positions per their maximum backtracable ply}
\label{count_cant_reach}
\hbox to\hsize{\hfil
\scalebox{1.2}{
   \begin{tabular}{|l|r|} \hline
       Ply & Number of Positions \\ \hline \hline
       0 & 6,650,818 \\ \hline
       1 & 4,175  \\ \hline
       2 & 2,494 \\ \hline
       3 & 114 \\ \hline
       4 & 104 \\ \hline
       5 & 2 \\ \hline
       6 & 4 \\ \hline
       7 & 1 \\ \hline
       8 & 1 \\ \hline
   \end{tabular}
}
\hfil}
\end{table}

\section{Conclusion and Future Work}
In this study, we estimated the number of reachable positions in Minishogi with high accuracy by creating candidate positions using uniform random numbers and measuring the proportion of those that can be reached through legal moves to the initial position. As a result, we found that the number of reachable positions is between $2.376\ldots \times 10^{18}$ and $2.379\ldots \times 10^{18}$ with a $95\%$ confidence interval. This result suggests that the number of reachable Minishogi positions is close to the upper bound found in previous research, making it challenging to strongly solve Minishogi using standard computational resources.

Our method demonstrates the effectiveness of uniform random sampling combined with reachability analysis, which can be applied to other combinatorial games. The authors are currently applying this method to standard Shogi, and the results obtained are under submission. Our program is available at {\footnotesize \url{https://github.com/u-tokyo-gps-tanaka-lab/minishogi-position-ranking}}.

\begin{acknowledgment}
This work was supported by JSPS KAKENHI Grant No. JP24K15244.
\end{acknowledgment}

\bibliographystyle{plain}
\bibliography{arxiv}

\end{document}